\documentclass[letterpaper, 10 pt, conference]{ieeeconf}
\IEEEoverridecommandlockouts
\usepackage{amsmath,amssymb,amsfonts}
\usepackage{framed}
\usepackage{algorithmic}
\usepackage{graphicx}
\usepackage{textcomp}
\usepackage{xcolor}
\usepackage[ruled,linesnumbered,vlined]{algorithm2e}
\usepackage{verbatim}
\usepackage{tikz}
\usepackage{pgfplots}
\usepackage{siunitx}
\pgfplotsset{compat=1.16}
\SetArgSty{textnormal}
 
\title{\LARGE \bf
  Terrain-Aware Kinodynamic Planning with Efficiently Adaptive State Lattices for Mobile Robot Navigation in Off-Road Environments*
}

\author{Eric R. Damm$^{1}$, Jason M. Gregory$^{2}$, Eli S. Lancaster$^{2}$, Felix A. Sanchez$^{2}$, Daniel M. Sahu$^{2}$, Thomas M. Howard$^{1}$% <-this % stops a space
\thanks{*Research was sponsored by the DEVCOM Army Research Laboratory (ARL) and was accomplished under Cooperative Agreement Number W911NF-20-2-0106. The views and conclusions contained in this document are those of the authors and should not be interpreted as representing the official policies, either expressed or implied, of the Army Research Laboratory or the U.S. Government. The U.S. Government is authorized to reproduce and distribute reprints for Government purposes notwithstanding any copyright notation herein.}
\thanks{$^{1}$Eric R. Damm and Thomas M. Howard are with the Robotics and Artificial Intelligence Laboratory, Hajim School of Engineering and Applied Sciences,
  University of Rochester, Rochester, NY, USA
        {\tt\small edamm@ur.rochester.edu}}%
\thanks{$^{2}$Jason M. Gregory, Eli S. Lancaster, Felix A. Sanchez, and Daniel M. Sahu are with the DEVCOM Army Research Laboratory, Adelphi, MD, USA
}
}
\begin{document}

\begin{minipage}{\textwidth}
\vspace{10cm}
\begin{center}
\copyright 2023 IEEE.  Personal use of this material is permitted.  Permission from IEEE must be obtained for all other uses, in any current or future media, including reprinting/republishing this material for advertising or promotional purposes, creating new collective works, for resale or redistribution to servers or lists, or reuse of any copyrighted component of this work in other works.
\end{center}
\end{minipage}

\maketitle
\thispagestyle{empty}
\pagestyle{empty}
\begin{abstract}
To safely traverse non-flat terrain, robots must account for the influence of terrain shape in their planned motions.
Terrain-aware motion planners use an estimate of the vehicle roll and pitch as a function of pose, vehicle suspension, and ground elevation map to weigh the cost of edges in the search space.  
Encoding such information in a traditional two-dimensional cost map is limiting because it is unable to capture the influence of orientation on the roll and pitch estimates from sloped terrain.   
The research presented herein addresses this problem by encoding kinodynamic information in the edges of a recombinant motion planning search space based on the Efficiently Adaptive State Lattice (EASL).
This approach, which we describe as a Kinodynamic Efficiently Adaptive State Lattice (KEASL), differs from the prior representation in two ways.
First, this method uses a novel encoding of velocity and acceleration constraints and vehicle direction at expanded nodes in the motion planning graph.
Second, this approach describes additional steps for evaluating the roll, pitch, constraints, and velocities associated with poses along each edge during search in a manner that still enables the graph to remain recombinant.  
Velocities are computed using an iterative bidirectional method using Eulerian integration that more accurately estimates the duration of edges that are subject to terrain-dependent velocity limits. 
Real-world experiments on a Clearpath Robotics Warthog Unmanned Ground Vehicle were performed in a non-flat, unstructured environment.  
Results from 2093 planning queries from these experiments showed that KEASL provided a more efficient route than EASL in 83.72\% of cases when EASL plans were adjusted to satisfy terrain-dependent velocity constraints.
An analysis of relative runtimes and differences between planned routes is additionally presented.
These results reinforce the importance of considering kinodynamic constraints for motion planning in non-flat environments and illustrate how such information can be encoded in an adaptive recombinant motion planning search space.
\end{abstract}

\IEEEpeerreviewmaketitle

\section{INTRODUCTION}
\label{sec:introduction}

\begin{figure}[tb]
    \centering
    \includegraphics[width=.9\linewidth]{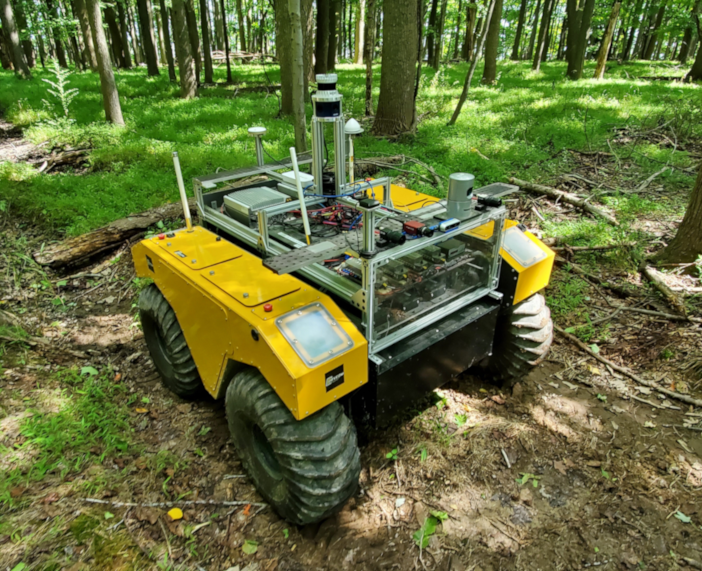}
    \caption{A Clearpath Robotics Warthog Unmanned Ground Vehicle in an off-road environment.  Kinodynamic Efficiently Adaptive State Lattices (KEASL) integrates velocity acceleration constraints imposed by terrain hazards into the search process to provide better global guidance in off-road environments.}
    \label{fig:warthog_in_gully}
\end{figure}
Mobile robot path planning for off-road environments is challenging because the terrain shape may create hazards that risk the safety and limit the performance of the vehicle.
In environments that exhibit both densely clustered obstacles and high slopes, the potential for vehicle rollover is high and should be accounted for by the robot's motion planner.
Traditionally, costs are integrated along the motions between nodes, known as edges, in a two-dimensional cost map where cost is approximately proportional to the safety of the vehicle at that position.
Such maps typically differentiate between traversable and non-traversable features, but are unable to capture the directional and velocity dependencies of risk associated with traversing these regions.
Examples of such features include hills, ditches, and fallen trees.
Incorporating this information into recombinant motion planning search spaces such as State Lattices \cite{dcmrmpsl} is difficult because the velocity at which a vehicle may arrive at a node is highly dependent on the route and any velocity constraints that may be imposed along those edges.
To address this, we reformulate the search process described by the Efficiently Adaptive State Lattice (EASL) \cite{impugv} which stores histories of vehicle poses, but does not account for the influence of roll, pitch, and velocity constraints - three important factors that should be considered when navigating through non-flat environments.
This approach, which we describe as the Kinodynamic Efficiently Adaptive State Lattice (KEASL), simulates vehicle motion at discretely sampled poses, assigns corresponding velocity constraints based on the estimated roll and pitch angles, and calculates the vehicle speed along the trajectory and the duration required to arrive at the current node.
The methodology presented provides a more physically accurate representation of edges in a recombinant motion planning search space than previous state lattice-based planners that neglect kinodynamic analysis or approximate it with two-dimensional cost maps.
In complex environments, the discretized sampling approach used by KEASL allows for a more accurate representation of the velocity dimension than the continuous approach used by EASL.
The main contributions of this research are the incorporation of kinodynamic analysis into state lattice-based motion planning, a novel representation of nodes that encode directionally dependent data (roll, pitch, velocities, and velocity constraints) in search, and methods for calculating vehicle velocities and velocity constraints within the context of this search space.
Experimental results presented in Section \ref{sec:results} illustrate that KEASL provided more efficient routes than EASL in 83.72\% of 2093 planning problems collected from real-world experiments on a Clearpath Robotics Warthog Unmanned Ground Vehicle when EASL plans were adjusted to satisfy terrain-dependent velocity constraints considered by KEASL.

\section{RELATED WORK}
\label{sec:related-work}

%\subsection{LOCAL PLANNING}
Mobile robots have most commonly considered the influence of terrain shape and kinodynamic constraints in trajectory generation and local motion planning. 
Kinodynamic modeling for model-predictive trajectory generation can be accomplished by the inversion of numerical predictive motion models \cite{orttgwmr}.
Trajectories are calculated using parametric optimal control techniques where velocity and/or curvature profiles are parameterized as a function of arclength.
Sampling-based methods for mobile robot navigation and path-following control \cite{mppi} naturally satisfy kinodynamic constraints because they are sampled in the vehicle's input space, however they provide less control over the distribution of sampled motions in the vehicle's state space.
Additional research has been performed in low friction environments where wheel slip was likely to occur.
These approaches utilized slip modeling and incorporated this analysis into a receding horizon control algorithm to alter the trajectory of the robot \cite{sampocpf}.
Receding horizon model-predictive control techniques considered the velocities of optimized motions, but did not consider the influence of roll and pitch of the vehicle and were subject to locally optimal solutions \cite{rhmpc}.

%\subsection{GLOBAL PLANNING}
In the context of global motion planning, consideration of kinodynamic constraints is less common because more simplified models of vehicle motion are required in order to achieve short planning times for long planning queries.  
State lattices \cite{dcmrmpsl,https://doi.org/10.1002/rob.20265} encodes differential constraints of vehicle motion using a precomputed edge library that naturally satisfies these constraints in a recombination motion planning search space.
The precomputed edges are calculated using speed-independent vehicle constraints and are stored in an edge library \cite{kmpslmp}.
As search occurs from start to goal, edges are removed from the graph if the motion crosses through a non-traversable cell in a cost map.
EASL \cite{impugv} introduced the idea of optimizing a discrete representation of node position and orientation based on the aggregate cost of edges to improve the relative optimality of planned motions.
One method for incorporating the influence of terrain shape in global planning appears in \cite{ompavtdt} where the gradient of the terrain is used to approximate roll and pitch calculations.
Other approaches such as \cite{emop} calculate the velocity of a robot at points along its planned trajectory.
This method outlined a three layer strategy where the end path duration was used as a core cost metric.
The researchers only considered flat environments allowing further opportunities to incorporate velocity primitives to be used by planning, as well as to develop new methods for creating and using discretized velocity profiles through non-flat terrain.
Other approaches such as \cite{4209319} use physics-based simulators to evaluate kinodynamic constraints in a Rapidly-Exploring Random Tree \cite{844730}, however such search spaces do not exhibit the recombinant structure of State Lattices that makes search in cluttered environments efficient.

%\subsection{DISCRIMINATORS}
The approach described in this paper addresses the problem of regional motion planning in environments where the effects of terrain shape influence the planned motion.
Cost is encoded in the graph as a linear combination of risk and duration.
Risk is represented from an encoding of non-traversable hazards in a two-dimensional cost map and by identifying estimated roll and pitch values that exceed tip-over constraints.
Duration is calculated as the total traversal time from the start state to any node in the graph.
This state lattice-based algorithm uses a novel representation of nodes that allows kinodynamic motion to be encoded in the recombinant graph.
The direction of motion at each node is encoded as forward, backward, or zero velocity.
This is preferable to the continuous representation of velocity in EASL which requires denser sampling in a continuous space.
The estimated roll and pitch values associated with poses from the path that connects the node to the node's backpointer is stored at each node.
The history of velocities and velocity constraints from the start state to the current node is additionally represented at each node.
Duration, which indicates the arrival time at a node relative to the time at the starting state, is calculated at each node as a function of the sampled velocities and lengths of the path segments from the start node to the current node.  
The method described herein results in higher fidelity motions that satisfy safety constraints that lower level path following controllers would otherwise have to compensate for.
Section \ref{sec:technical-approach} goes into detail on the unique construction of the search space, and the definition of a node in this context.
Section \ref{sec:experimental-design} discusses the tests that were performed, why they were chosen, and how they demonstrate the value of kinodynamic aware motion planning.
Finally, Sections \ref{sec:results} and \ref{sec:discussion} illustrate the relative performance of KEASL and EASL from real-world experiments and demonstrate the significance that kinodynamic consideration has on robot safety and performance for mobile robot path planning in non-flat environments. 
\section{TECHNICAL APPROACH}
\label{sec:technical-approach}

\begin{figure}[htb]
  \centering
  \vspace{.1in}
  \begin{tikzpicture}[textnode/.style={anchor=mid,font=\tiny},nodeopen/.style={circle,draw=black!80,fill=white,minimum size=4mm,font=\tiny},nodeclosed/.style={circle,draw=black!80,fill=black!20,minimum size=4mm,font=\tiny,top color=white,bottom color=black!40},nodeunexplored/.style={circle,dashed,draw=black!80,fill=white,minimum size=4mm,font=\tiny}]
  	\draw[->] (0,1) arc (270:345:1);
	\draw[->] (0,1) arc (-270:-345:1);
	\draw[->] (1,1) arc (270:345:1);
	\draw[->] (1,1) arc (-270:-345:1);
	\draw[->] (2,1) arc (270:345:1);
	\draw[->] (2,1) arc (-270:-345:1);
	\draw[->] (3,1) arc (270:345:1);
	\draw[->] (3,1) arc (-270:-345:1);
	\draw[->] (4,1) arc (270:345:1);
	\draw[->] (4,1) arc (-270:-345:1);
	\draw[->] (3,0) arc (0:75:1);
	\draw[->] (3,0) arc (180:105:1);
	\draw[->] (1,0) arc (180:255:1);
	\draw[->] (1,0) arc (0:-90:1);
	\draw[->] (2,-1) arc (270:345:1);
	\draw[->] (2,-1) arc (-270:-345:1);
	\node[nodeunexplored] (nodeunexplored1) at (0,2) {};
	\node[nodeopen] (nodea) at (1,2) {A};
	\node[nodeopen] (nodeb) at (2,2) {B};
	\node[nodeopen] (nodec) at (3,2) {C};
	\node[nodeopen] (noded) at (4,2) {D};
	\node[nodeopen] (nodee) at (5,2) {E};
	\node[nodeclosed] (nodef) at (0,1) {F};
	\node[nodeclosed] (nodeg) at (1,1) {G};
	\node[nodeclosed] (nodeh) at (2,1) {H};
	\node[nodeclosed] (nodei) at (3,1) {I};
	\node[nodeclosed] (nodej) at (4,1) {J};
	\node[nodeopen] (nodek) at (5,1) {K};
	\node[nodeunexplored] (nodeunexplored2) at (0,0) {};
	\node[nodeclosed] (nodel) at (1,0) {L};
	\node[nodeopen] (nodem) at (2,0) {M};
	\node[nodeclosed] (noden) at (3,0) {N};
	\node[nodeopen] (nodeo) at (4,0) {O};
	\node[nodeopen] (nodeo) at (5,0) {P};
	\node[nodeopen] (nodep) at (0,-1) {Q};
	\node[nodeopen] (nodeq) at (1,-1) {R};
	\node[nodeclosed] (noder) at (2,-1) {S};
	\node[nodeopen] (nodes) at (3,-1) {T};
	\node[nodeunexplored] (nodeunexplored3) at (4,-1) {};
	\node[nodeunexplored] (nodeunexplored4) at (5,-1) {};
	\node[nodeunexplored] (nodeunexplored5) at (0,-2) {};
	\node[nodeunexplored] (nodeunexplored6) at (1,-2) {};
	\node[nodeunexplored] (nodeunexplored7) at (2,-2) {};
	\node[nodeopen] (nodex) at (3,-2) {U};
	\node[nodeunexplored] (nodeunexplored8) at (4,-2) {};
	\node[nodeunexplored] (nodeunexplored9) at (5,-2) {};
	\node[nodeclosed] (nodeclosed) at (7,1) {};
	\node[nodeopen] (nodeopen) at (7,1.5) {};
	\node[nodeunexplored] (nodeunexplored) at (7,2) {};
	\node[textnode] (unexploredlabel) at (6.25,2) {unexplored};
	\node[textnode] (openlabel) at (6.25,1.5) {open};
	\node[textnode] (closedlabel) at (6.25,1) {closed};
	\node[textnode] (edgelabel) at (6.25,0.25) {edge};
	\draw[->] (6.5,0) arc (270:345:1);
    \draw[->] (nodef) to (nodeg);
    \draw[->] (nodeg) to (nodeh);
    \draw[->] (nodeh) to (nodei);
    \draw[->] (nodel) to (nodeq);
    \draw[->] (noder) to (nodes);
    \draw[->] (noden) to (nodei);
    \draw[->] (nodei) to (nodej);
    \draw[->] (nodej) to (nodek);
    \draw[->] (6.25,-2.25) to (6.25,-1);
    \draw[->] (6,-2) to (7.25,-2);
    \node[textnode] (plusxlabel) at (7.5,-2) {+x};
    \node[textnode] (plusylabel) at (6.25,-0.75) {+y};
    \node[textnode] (plusyawlabel) at (6.875,-1.5) {+$\psi$};
    \draw[->] (6.62,-2.303) arc (-45:135:0.5);
  \end{tikzpicture}  
  \caption{Illustration of seven KEASL edge library expansions of a state lattice indexed by $\left(x,y,\psi\right)$ starting from node $\left(F,0^{\circ}\right)$.  Expansions of the state lattice are performed at $\left(L,270^{\circ}\right)$, $\left(S,0^{\circ}\right)$, $\left(N,90^{\circ}\right)$, $\left(G,0^{\circ}\right)$, $\left(H,0^{\circ}\right)$, $\left(I,0^{\circ}\right)$ and $\left(J,0^{\circ}\right)$.}
  \label{fig:lattice-example}
\end{figure}
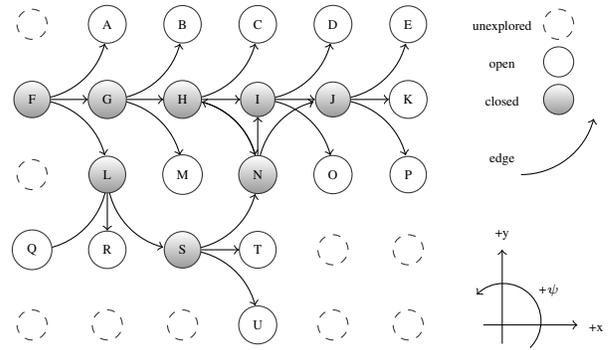

Consider the illustration of the search space depicted in Figure \ref{fig:lattice-example}.
This figure represents nodes explored during the search as circles, where the shading of the circle indicates whether it is currently on the open list, the closed list, or has not yet been encountered by the search process.
For illustrative purposes, the position of each node is labeled with a letter, but the position alone does not fully specify a node in this graph.
Each node is defined with a discrete representation of the $x,y$ position, a discretized heading ($\psi$), and direction $d$ that indicates whether the vehicle is moving forwards ($+$), backwards ($-$), or is stopped ($0$). 
Search begins from node F at heading $\psi = 0^{\circ}$ and direction $d=0$, which we denote as $\left(F,0^{\circ},0\right)$, towards a goal in the distance in the positive $x$ direction.
The first expansion of the edge library expands node $\left(F,0^{\circ},0\right)$ to nodes $\left(A,90^{\circ},+\right)$, $\left(G,0^{\circ},+\right)$, and $\left(L,270^{\circ},+\right)$.
The paths $p_{F_{0^{\circ},0}A_{90^{\circ},+}}$, $p_{F_{0^{\circ},0}G_{0^{\circ},+}}$, and $p_{F_{0^{\circ}}L_{270^{\circ},+}}$ represent the position, orientation, and curvature of each path as defined by the precomputed edge library.
Additional details on how precomputed paths are loaded from the edge library is provided in Section \ref{sec:technical-approach-paths}.
The attitude and elevation of the robot at each pose along each path is represented as $\alpha_{F_{0^{\circ},0}A_{90^{\circ},+}}$, $\alpha_{F_{0^{\circ},0}G_{0^{\circ},+}}$, and $\alpha_{F_{0^{\circ},0}L_{270^{\circ},+}}$ respectively.
Because the attitude and elevation are specific to the vehicle pose with respect to a predefined or sensed ground elevation map, these need to be computed for each edge library expansion.
Additional details on attitude elevation profiles is provided in Section \ref{sec:technical-approach-attitude-elevation-profiles}.
The velocity acceleration constraint profiles $\beta_{F_{0^{\circ},0}A_{90^{\circ},+}}$, $\beta_{F_{0^{\circ},0}G_{0^{\circ},+}}$, and $\beta_{F_{0^{\circ},0}L_{270^{\circ},+}}$ represent the velocity acceleration constraints from node $\left(F,0^{\circ},0\right)$ to nodes $\left(A,90^{\circ},+\right)$, $\left(G,0^{\circ},+\right)$, and $\left(L,270^{\circ},+\right)$ respectively.
Velocity acceleration constraints are computed at each pose along each expanded edge.
Since each of these edges start with $d=0$ and terminates with $d=+$, the defaults for the minimum velocity and minimum acceleration are $0 \frac{m}{sec}$ and $0 \frac{m}{sec^{2}}$ respectively.
The maximum velocity and acceleration are defined by the vehicle limits, curvature, attitude, or any other externally defined condition that may limit the vehicle velocity at that pose.
Additional details on velocity acceleration constraint profiles is provided in Section \ref{sec:technical-approach-velocity-acceleration-constraint-profiles}.
The velocity acceleration profiles $\gamma_{F_{0^{\circ},0}A_{90^{\circ},+}}$, $\gamma_{F_{0^{\circ},0}G_{0^{\circ},+}}$, and $\gamma_{F_{0^{\circ},0}L_{270^{\circ},+}}$ represent the actual velocities at each pose that satisfy the velocity acceleration constraint profiles from node $\left(F,0^{\circ},0\right)$ to nodes $\left(A,90^{\circ},+\right)$, $\left(G,0^{\circ},+\right)$, and $\left(L,270^{\circ},+\right)$.

After the first expansion, the node $\left(L,270^{\circ},+\right)$ expands to nodes $\left(Q,180^{\circ},+\right)$, $\left(R,270^{\circ},+\right)$, and $\left(S,0^{\circ},+\right)$.
Paths $p_{L_{270^{\circ},+}Q_{180^{\circ},+}}$, $p_{L_{270^{\circ},+}R_{270^{\circ},+}}$, and $p_{L_{270^{\circ},+}S_{0^{\circ},+}}$ are applied from the precomputed edge library and attitude elevation profiles $\alpha_{L_{270^{\circ},+}Q_{180^{\circ},+}}$, $\alpha_{L_{270^{\circ},+}R_{270^{\circ},+}}$, and $\alpha_{L_{270^{\circ},+}S_{0^{\circ},+}}$ are calculated for each of these new poses.
In addition to the representation of attitude elevation profiles in the search space, KEASL deviates from the graph representation in EASL by explicitly representing the velocity acceleration constraints along each edge.
If velocity acceleration constraints are only a function of the instantaneous pose of the vehicle, then these velocity acceleration constraint profiles can be represented as $\beta_{L_{270^{\circ},+}Q_{180^{\circ},+}}$, $\beta_{L_{270^{\circ},+}R_{270^{\circ},+}}$, and $\beta_{L_{270^{\circ},+}S_{0^{\circ},+}}$ since $\beta_{F_{0^{\circ},0}L_{270^{\circ},+}}$ will be unaffected by the motion from the differences in the motions from node $\left(L,270^{\circ},+\right)$ to nodes $\left(Q,180^{\circ},+\right)$, $\left(R,270^{\circ},+\right)$, and $\left(S,0^{\circ},+\right)$.
If however the velocity acceleration constraints are a function of the attitude or elevation rates of the vehicle, as may occur when traversing undulations in the terrain or negotiating over fallen trees, the full profile of velocity acceleration constraints $\beta_{F_{0^{\circ},0}L_{270^{\circ},+}Q_{180^{\circ},+}}$, $\beta_{F_{0^{\circ},0}L_{270^{\circ},+}R_{270^{\circ},+}}$, and $\beta_{F_{0^{\circ},0}L_{270^{\circ},+}S_{0^{\circ},+}}$ must be represented as it is a function of the transition between paths $p$ and attitude elevation profiles $\alpha$ of neighboring edges.  
KEASL additionally represents the profile of velocities and accelerations from the start node to the current, which for these nodes are $\gamma_{F_{0^{\circ},0}L_{270^{\circ},+}Q_{180^{\circ},+}}$, $\gamma_{F_{0^{\circ},0}L_{270^{\circ},+}R_{270^{\circ},+}}$, and $\gamma_{F_{0^{\circ},0}L_{270^{\circ},+}S_{0^{\circ},+}}$.
The reason why this is necessary is evident from the two routes that arrive at $\left(K,0^{\circ},+\right)$.  
The first route passes from node $\left(F,0^{\circ},0\right)$ to node $\left(K,0^{\circ},+\right)$ through nodes $\left(G,0^{\circ},+\right)$, $\left(H,0^{\circ},+\right)$, $\left(I,0^{\circ},+\right)$, and $\left(J,0^{\circ},+\right)$.
The second route passes from node $\left(F,0^{\circ},0\right)$ to node $\left(K,0^{\circ},+\right)$ through nodes $\left(L,270^{\circ},+\right)$, $\left(S,0^{\circ},+\right)$, $\left(N,90^{\circ},+\right)$, and $\left(J,0^{\circ},+\right)$.
The first route arrives at node $\left(K,0^{\circ},+\right)$ with velocity acceleration constraint profile $\beta_{F_{0^{\circ},0}G_{0^{\circ},+}H_{0^{\circ},+}I_{0^{\circ},+}J_{0^{\circ},+}K_{0^{\circ},+}}$ and velocity acceleration profile $\gamma_{F_{0^{\circ},0}G_{0^{\circ},+}H_{0^{\circ},+}I_{0^{\circ},+}J_{0^{\circ},+}K_{0^{\circ},+}}$ whereas the second route arrives at node $\left(K,0^{\circ},+\right)$ with velocity acceleration constraint profile $\beta_{F_{0^{\circ},0}L_{270^{\circ},+}S_{0^{\circ},+}N_{90^{\circ},+}J_{0^{\circ},+}K_{0^{\circ},+}}$ and velocity acceleration profile $\gamma_{F_{0^{\circ},0}L_{270^{\circ},+}S_{0^{\circ},+}N_{90^{\circ},+}J_{0^{\circ},+}K_{0^{\circ},+}}$.
If the velocity acceleration constraints between nodes $\left(I,0^{\circ},+\right)$ and $\left(J,0^{\circ},+\right)$ are less conservative than the velocity acceleration constraints between nodes $\left(N,90^{\circ},+\right)$ and $\left(J,0^{\circ},+\right)$, then each route may arrive at node $J$ with a different linear and angular velocity.
Since the duration instead of distance is a more useful metric to use when deciding the optimality of paths in a kinodynamic planner, we recognize that the duration of each motion can be calculated through the integration of these linear velocities along each distance parameterized motion.
The duration calculation is shown in Algorithm \ref{alg:path-duration}.
Although representing the full velocity acceleration constraint profile and velocity acceleration profile at each node requires more storage and operations than the approach described in EASL, it more effectively captures the impact of velocity constraints that may be a function of both the path and the environment.

\subsection{Paths}
\label{sec:technical-approach-paths}
Paths in each edge are represented by a vector of positions, orientations, and steering curvatures.
The shape of the path instantiated at each edge is interpolated from a two-dimensional lookup table parameterized by relative position and heading using the techniques described in \cite{orttgwmr}.
Interpolation is necessary to accommodate the local adaptation of node position and orientation indices from the node optimization procedure described in \cite{impugv}.  
Intersections with non-traversable hazards encoded in a two-dimensional cost map are evaluated using a method based on convolving the vehicle footprint as described in \cite{dcmrmpsl} and \cite{impugv}.  
Whereas EASL would then take the length of this path and the boundary velocities encoded at each node to estimate the duration of the edge, KEASL estimates the roll, pitch, velocity acceleration constraints, and velocity profile to provide a more accurate prediction of the duration of each edge.
Subsections \ref{sec:technical-approach-attitude-elevation-profiles}, \ref{sec:technical-approach-velocity-acceleration-constraint-profiles}, and \ref{sec:technical-approach-velocity-acceleration-profiles} outline this procedure.

\subsection{Attitude Elevation Profiles}
\label{sec:technical-approach-attitude-elevation-profiles}

In the KEASL search space, an attitude elevation profile is a list of the roll, pitch, linear displacement, angular displacement, observation status, and elevation values for the vehicle states between a node and its parent.
The roll and pitch values are calculated using a quasistatic suspension model that makes it a function of the current pose, the robot kinematics, and the estimated or prior height map.
This makes the calculation of roll and pitch values independent of past and future poses, however the dependence on the height map eliminates the option of precomputing these values.
Because of this, each node only needs to store the attitude elevation profile that begins from the node's parent.
Edges that exhibit roll and pitch values that exceed vehicle-defined limits will be marked as non-traversable and eliminated as options during search.
This information is also used to set velocity acceleration constraints based on roll and pitch values, as well as for unobserved areas of the height map.

The method for estimating the roll and pitch angles of the vehicle in Section \ref{sec:results} used a method based on quasi-static analysis that assumed contact at three or more wheels for a four-wheeled unmanned ground vehicle.
At each state, the height map was sampled at each of the vehicle's wheel locations.
Bi-linear interpolation of the height map cells was used to achieve smoother wheel contact point estimates.
The location of the robot's center of gravity and wheel heights were then used to determine three wheel contact points.
Once the three contact points were known, the vehicle's roll, pitch, and elevation values were calculated.
After these initial roll and pitch values were found, a suspension model was used to update the attitude estimation and better simulate how the vehicle would be oriented.
These estimated roll and pitch angles were then added to the attitude elevation profile.

\subsection{Velocity Acceleration Constraint Profiles}
\label{sec:technical-approach-velocity-acceleration-constraint-profiles}

The velocity acceleration constraint profile contains information about the maximum allowable velocity and acceleration values at points along expanded edges.
Unlike the attitude elevation information, future calculations may impact prior constraint values.
For example, if the attitude elevation information between node I to node J in Figure \ref{fig:lattice-example} indicates the need for a velocity constraint between nodes H and I, this constraint is unique to the path through nodes H, I, and J.
If instead the path was through nodes H, I and O, there may not be a constraint set between nodes H and I.
In this scenario the future expansions impact the prior profiles and there would be two unique velocity acceleration constraint profiles stored in node I, even though all its children were identical.
The constraints were calculated based on the robot's dimensions and location of the center of gravity.
At roll values closest to the tip-over point of the robot, only the lowest velocities are allowed.
This velocity limit is increased as the roll decreases and the roll risk decreases.
The velocity limits from pitch are determined from the risk of vehicle maneuvers when traversing hills.
When a vehicle is going uphill, it is important that there is full commitment to the maneuver to give the best chances of success.
If an uphill maneuver is aborted partway through, there is an increased risk for unexpected behavior like wheel slip and vehicle tip-over.
Because of this, the velocity is only limited at extreme uphill (negative) pitches to give the best chance of hill traversal.
The opposite is the case when navigating down a hill.
The safest way to perform this action is to point directly downhill and maintain a slow velocity while also maintaining grip of all four wheels.
This expectation is captured in the downhill (positive) pitch velocity constraints where there is a sharp drop in velocity allowance starting at low pitch values.
It follows that there is also a continually decreasing limit as the pitch value increases.
A plot of the attitude-based velocity constraints being used for the experimental results in Section \ref{sec:results} is shown in Figure \ref{fig:attitude_vel_constraints}.
Velocity constraints are also set for areas of the height map that are unobserved, to help ensure safe traversal through unseen sections of the environment.

\subsection{Velocity Acceleration Profiles}
\label{sec:technical-approach-velocity-acceleration-profiles}
\begin{figure}[htb]
  \centering
  \resizebox{.9\columnwidth}{!}{
  \begin{tikzpicture}[textnode/.style={anchor=mid,font=\tiny}]
  	\draw[->] (0,5.75) to (0,8);
    \draw[->] (-0.25,6) to (7,6);
    \draw[-,dashed] (0,7) to (1,7);
    \draw[-,dashed] (1,7) to (1,7.5);
    \draw[-,dashed] (1,7.5) to (2,7.5);
    \draw[-,dashed] (2,7.5) to (2,6.5);
    \draw[-,dashed] (2,6.5) to (4,6.5);
    \draw[-,dashed] (4,6.5) to (4,8);
    \draw[-,dashed] (4,8) to (6,8);
    \draw[-,dashed] (6,8) to (6,7);
    \draw[-,dashed] (6,7) to (7,7);
    \node[textnode] (timelabel) at (3.5,5.75) {time $\left(sec\right)$};
    \node[font=\footnotesize] (alabel) at (3.5,5.375) {(a)};
    \node[textnode] (velocitylabel) at (-0.625,7) {velocity};
    \node[textnode] (velocityunitlabel) at (-0.625,6.625) {$\left(\frac{m}{sec}\right)$};
    \draw[->] (0,2.75) to (0,5);
    \draw[->] (-0.25,3) to (7,3);
    \draw[-,dashed] (0,4) to (1,4);
    \draw[-,dashed] (1,4) to (1,4.5);
    \draw[-,dashed] (1,4.5) to (2,4.5);
    \draw[-,dashed] (2,4.5) to (2,3.5);
    \draw[-,dashed] (2,3.5) to (4,3.5);
    \draw[-,dashed] (4,3.5) to (4,5);
    \draw[-,dashed] (4,5) to (6,5);
    \draw[-,dashed] (6,5) to (6,4);
    \draw[-,dashed] (6,4) to (7,4);
    \draw[-,color=red] (0,3) to (0.5,4);
    \draw[-,color=red] (0.5,4) to (1.0,4);
    \draw[-,color=red] (1.0,4) to (1.25,4.5);
    \draw[-,color=red] (1.25,4.5) to (2.0,4.5);
    \draw[-,color=red] (2.0,4.5) to (2.0,3.5);
    \draw[-,color=red] (2.0,3.5) to (4.0,3.5);
    \draw[-,color=red] (4.0,3.5) to (4.75,5);
    \draw[-,color=red] (4.75,5) to (6,5);
    \draw[-,color=red] (6,5) to (6,4);
    \draw[-,color=red] (6,4) to (7,4);
    \node[textnode] (timelabel) at (3.5,2.75) {time $\left(sec\right)$};
    \node[font=\footnotesize] (blabel) at (3.5,2.375) {(b)};
    \node[textnode] (velocitylabel) at (-0.625,4) {velocity};
    \node[textnode] (velocityunitlabel) at (-0.625,3.625) {$\left(\frac{m}{sec}\right)$};
    \draw[->] (0,-0.25) to (0,2);
    \draw[->] (-0.25,0) to (7,0);
    \draw[-,dashed] (0,1) to (1,1);
    \draw[-,dashed] (1,1) to (1,1.5);
    \draw[-,dashed] (1,1.5) to (2,1.5);
    \draw[-,dashed] (2,1.5) to (2,0.5);
    \draw[-,dashed] (2,0.5) to (4,0.5);
    \draw[-,dashed] (4,0.5) to (4,2);
    \draw[-,dashed] (4,2) to (6,2);
    \draw[-,dashed] (6,2) to (6,1);
    \draw[-,dashed] (6,1) to (7,1);
    \draw[-,color=red] (0,0) to (0.5,1);
    \draw[-,color=red] (0.5,1) to (1.0,1);
    \draw[-,color=red] (1.0,1) to (1.25,1.5);
    \draw[-,color=red] (1.25,1.5) to (2.0,1.5);
    \draw[-,color=red] (2.0,1.5) to (2.0,0.5);
    \draw[-,color=red] (2.0,0.5) to (4.0,0.5);
    \draw[-,color=red] (4.0,0.5) to (4.75,2);
    \draw[-,color=red] (4.75,2) to (6,2);
    \draw[-,color=red] (6,2) to (6,1);
    \draw[-,color=red] (6,1) to (7,1);
    \draw[-,color=blue] (0,0) to (0.5,1);
    \draw[-,color=blue] (0.5,1) to (1.0,1);
    \draw[-,color=blue] (1.0,1) to (1.25,1.5);
    \draw[-,color=blue] (1.25,1.5) to (1.5,1.5);
    \draw[-,color=blue] (1.5,1.5) to (2.0,0.5);
    \draw[-,color=blue] (2.0,0.5) to (4.0,0.5);
    \draw[-,color=blue] (4.0,0.5) to (4.75,2);
    \draw[-,color=blue] (4.75,2) to (5.5,2);
    \draw[-,color=blue] (5.5,2) to (6,1);
    \draw[-,color=blue] (6,1) to (7,1);
    \node[textnode] (timelabel) at (3.5,-0.25) {time $\left(sec\right)$};
    \node[font=\footnotesize] (clabel) at (3.5,-0.625) {(c)};
    \node[textnode] (velocitylabel) at (-0.625,1) {velocity};
    \node[textnode] (velocityunitlabel) at (-0.625,0.625) {$\left(\frac{m}{sec}\right)$};
  \end{tikzpicture}
  }
  \caption{A depiction of the steps in the bi-directional Eulerian integration approach used for velocity acceleration profile calculation. (a) depicts the velocity constraints along an example trajectory. (b) shows the forward pass, where the calculations occur from left to right and never exceed the constraint values. (c) illustrates the backward pass where calculations occur from right to left and never exceed the forward pass values. The resulting shape is the velocity profile for the path.}
  \label{fig:nonflat5_fb}
\end{figure}
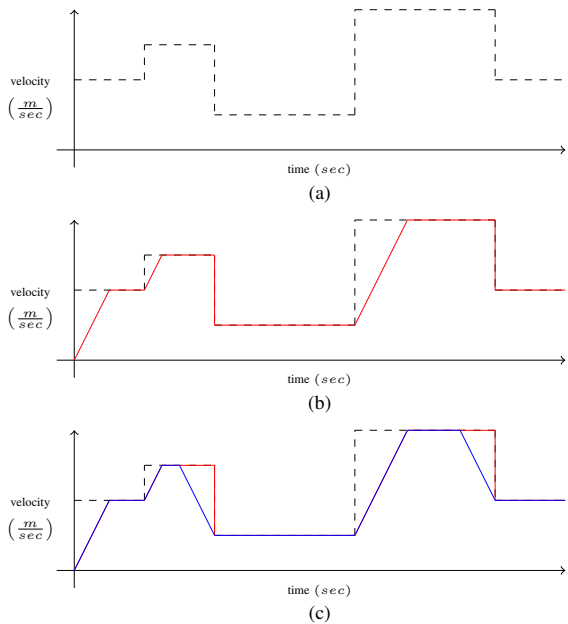

The velocity acceleration profiles contain information about the linear velocity, angular velocity, linear acceleration, angular acceleration, and time at states along a trajectory.
They are created via a bidirectional Eulerian integration method which depends on the velocity acceleration constraint profile and the distances between sampled points.
Figure \ref{fig:nonflat5_fb} depicts the process by which this approach is performed.
Plot (a) shows the velocity constraints along an example trajectory.
Plot (b) shows the first, forward pass.
From the first point, velocity is increasing at the maximum acceleration limit.
Along the path, there are velocity limits determined from things like projected roll, pitch, curvature limits, and acceleration limits. 
At each of these velocity limits along the path, the current velocity (the one that has been constantly accelerating) is dropped down to the limit and the acceleration begins to the next points.
Plot (c) shows the final step of the backward pass through the profile.
The process for this step is similar to that in (b) except this process runs from the last point backward to the first point.
The boundary condition for the backward pass is the profile created by the forward pass instead of just the velocity constraints.
The methodology for the approach is further described in Algorithm \ref{alg:forward-backward-velocity-search}, and the duration calculation is shown in Algorithm \ref{alg:path-duration}.
The combination of accurate velocity profiling with exact time values leads to more accurate predictions on the total duration of the robot's motion.

Comprehensive histories of these calculations are stored at each node because, similarly to the velocity acceleration constraint profiles discussed in Section \ref{sec:technical-approach-velocity-acceleration-constraint-profiles}, each of these values may be impacted by past and/or future values.
%
\begin{comment}
Referring to Figure \ref{fig:lattice-example}, if there is a velocity constraint between nodes I and J, the vehicle may need to begin slowing down between nodes H and I in order to meet that constraint.
%
If instead a path between nodes I and O is chosen, and no slowdown is needed between H and I, there are two unique velocity acceleration profiles between nodes H and I.
%
There is one for the path containing nodes H, I, and O where the vehicle is at a constant top speed, and one for path containing nodes H, I, and J where a deceleration begins to occur.
%
\end{comment}
Because of this, it is necessary to encode velocity constraint information throughout the search space as unique profiles for full histories rather than just for edges between nodes.

\begin{algorithm}\small{
\caption{\sc Discretized Velocity Profile} \label{alg:forward-backward-velocity-search}
\KwIn{Velocity\_Acceleration\_Constraints $\boldsymbol{\beta}$, Velocities\_Accelerations $\boldsymbol{\gamma}$, Linear\_Distances\_Between\_Points $\textbf{ld}$}
\KwOut{Velocities\_Accelerations $\boldsymbol{\gamma}_{1:n}$ for $\textbf{n}=$\sc Length$(\boldsymbol{\beta})$}
\SetKwProg{Fn}{Function}{:}{}
\SetKwFunction{FForwardBackwardPass}{\sc ForwardBackwardPass(\textnormal{$\boldsymbol{\beta}$, $\boldsymbol{\gamma}$, $\textbf{ld}$})}
\Fn{\FForwardBackwardPass}{
  \DontPrintSemicolon
  $\textbf{i}=1$ \\
  \While{$\textbf{i} < \textbf{n}$} {
    $\boldsymbol{\gamma}_{vel_{i+1}} = \sqrt{\boldsymbol{\gamma}_{vel_{i}}^2 + 2*\boldsymbol{\gamma}_{acc_{i}}*\textbf{ld}_i}$ \\
    \If{$\boldsymbol{\gamma}_{vel_{i+1}} > \boldsymbol{\beta}_{vel_{i+1}}$} {
      $\boldsymbol{\gamma}_{vel_{i+1}} = \boldsymbol{\beta}_{vel_{i+1}}$
    }
    $\textbf{i} \gets \textbf{i}+1$
  }
  $\textbf{i}=n$ \\
  \While{$\textbf{i} > 1$} {
    $\textbf{temp\_vel} = \sqrt{\boldsymbol{\gamma}_{vel_{i}}^2 + 2*\boldsymbol{\gamma}_{acc_{i}}*\textbf{ld}_{i-1}}$ \\
    \If{$\textbf{temp\_vel} < \boldsymbol{\gamma}_{vel_{i-1}}$} {
      $\boldsymbol{\gamma}_{vel_{i-1}} = \textbf{temp\_vel}$
    }
    $\textbf{i} \gets \textbf{i}-1$
  }
  \Return {$\boldsymbol{\gamma}$}\;
}
}
\end{algorithm}

\begin{algorithm}\small{
\caption{\sc Path Duration} \label{alg:path-duration}
\KwIn{Velocities\_Accelerations $\boldsymbol{\gamma}$}
\KwOut{Duration $\textbf{D}_{1:n}$}
\SetKwProg{Fn}{Function}{:}{}
\SetKwFunction{FDuration}{\sc Duration(\textnormal{$\boldsymbol{\gamma}$})}
\Fn{\FDuration}{
  \DontPrintSemicolon
  $\textbf{i}=1$ \\
  $\textbf{D}=0$ \\
  \While{$\textbf{i} < \textbf{n}$} {
    $\textbf{t}_{temp} = \frac{|\boldsymbol{\gamma}_{vel_{i+1}} - \boldsymbol{\gamma}_{vel_{i}}|}{|\boldsymbol{\gamma}_{accel_{i}}|}$ \\
    $\textbf{D} \gets \textbf{D}+\textbf{t}_{temp}$ \\
    $\textbf{i} \gets \textbf{i}+1$
  }
  \Return {$\textbf{D}$}\;
}
}
\end{algorithm}
\section{EXPERIMENTAL DESIGN}
\label{sec:experimental-design}

\begin{figure}
    \centering
    \resizebox{.95\columnwidth}{!}{\input{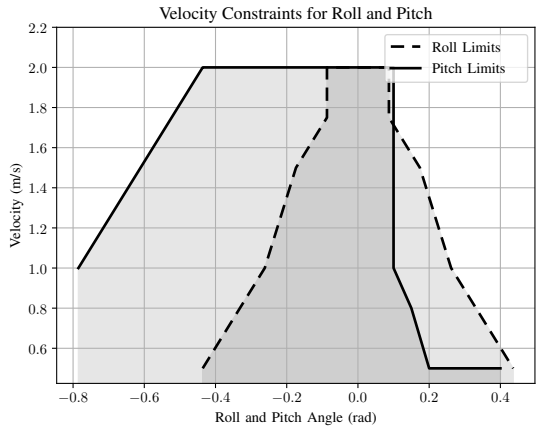}}
    \caption{Velocity limits set by the roll and pitch constraints. These indicate the operating ranges that we deem ``safe'' for robot traversal for these experiments. Anything exceeding these limits is considered unsafe. The operating range for roll and pitch is indicated by the shaded gray area beneath each line.}
    \label{fig:attitude_vel_constraints}
\end{figure}

Experiments were designed to compare the planning time, path duration, and satisfaction of roll, pitch, and velocity constraints of EASL and KEASL across a variety of environments.
Cost and height maps for planning problems were collected from real-world experiments on a Clearpath Robotics Warthog Unmanned Ground Vehicle in non-flat terrain.
The velocity constraints used in these physical experiments are depicted in Figure \ref{fig:attitude_vel_constraints}.
Velocity constraints are computed as a function of the estimated roll and pitch values at each robot pose.
Speeds are limited in a manner that is approximately proportional to the positive pitch angle and magnitude of the roll angle to slow the vehicle down when it is travelling downhill or on significant side slopes.
Deviation from the current direction of downwards momentum in these scenarios at speed can lead to unrecoverable vehicle behavior such as over steer or tip-overs.
The velocity constraints set by high chassis roll angles are also very strict because of the roll-over risk.
Negative pitch values penalize velocity limits the least because at high slopes, momentum must be carried in order to successfully traverse the feature.
If the velocity limit is set too low and the vehicle fails to complete the maneuver to the top of a hill, unsafe behavior can ensue.
At too steep of a pitch, a vehicle may be unable to turn around safely (or at all) to descend and either retry the ascent or replan a different route.
Additional velocity constraints, such as those that limit vehicle speed based on a combination of roll, pitch, and angular velocity, could be used in place of this representative model if the application warrants. 
For these real-world tests, the robot navigated through three locations with different terrain features.
During these physical experiments, new plans were periodically requested to a fixed goal from the robot's current pose.
For each of these planning queries, 24 unique goal states were sampled at a uniform distance from the robot to better capture the variety of potential routes through an environment.
If a goal was located on a lethal cost cell, the plan was discarded and not counted in the data analysis process.
Figure \ref{fig:EASL_multi_goal_1} shows examples of the 24 goal states and corresponding paths.

A maximum velocity of 2\si{m/s} and maximum acceleration of 2\si{m/s^{2}} was used for all experiments.
A 0.2 meter resolution cost map represented non-traversable hazards in the environment.
A similarly sampled ground elevation map stores the estimated height at each cell and an indication of whether the height was observed.
If a cell was marked as unobserved, the height value in the cell was ignored and a strict velocity constraint was applied to any edge running through it.
A maximum planning time of 10 seconds was used to ensure that both EASL and KEASL were able to reach a solution.
It should be noted that 10 seconds was far longer than was needed for both planning algorithms, and the majority of solutions were found in less than one second.

\begin{comment}
tent with non-flat terrain. The focus
of the work was to compare the outputs of KEASL and
EASL using a main evaluation metric of the relative safety
of the chosen path by determining if the robot violated roll
or velocity constraints along its path.
The velocity constraints set by roll and pitch limits are
depicted in Figure 5. These were chosen based on the dimen-
sions of the Warthog, and on the complexity of the obstacles
that were to be traversed. The most severe constraint was set
to positive pitches, because of the variance in expected mo-
tions duei
\end{comment}
\section{RESULTS}
\label{sec:results}

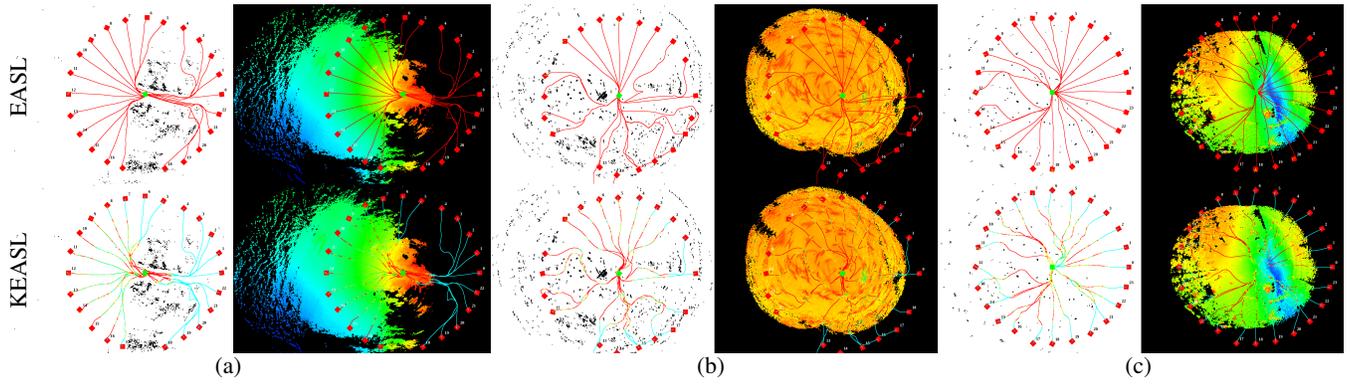
\begin{figure*}[htb]
    \centering
    \begin{tikzpicture}
        \node[anchor=south west,inner sep=0] (image) at (0,0) {\includegraphics[width=1\textwidth]{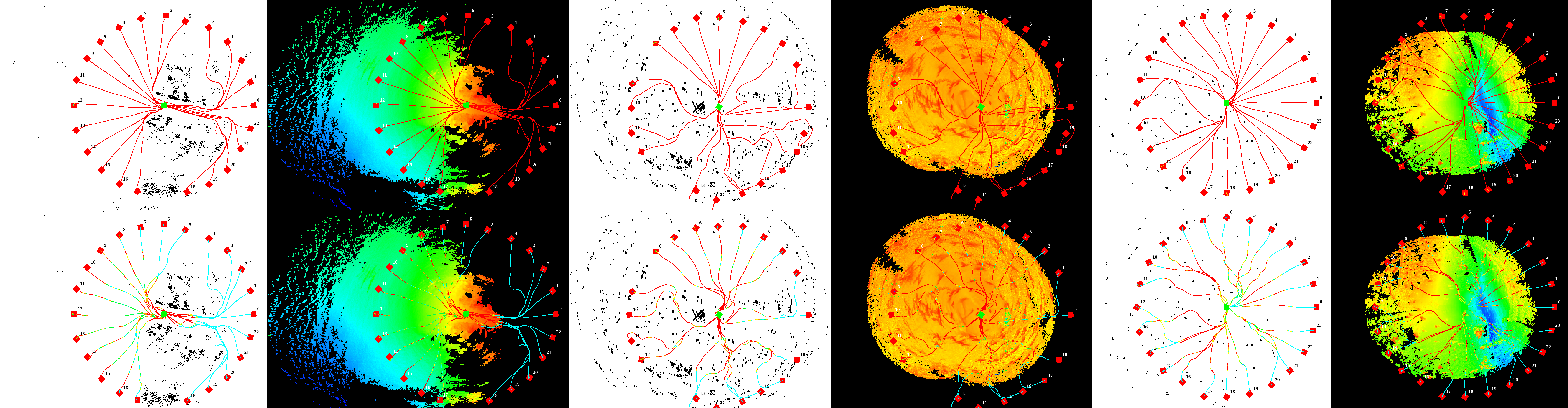}};

        \begin{scope}[x={(image.south east)},y={(image.north west)}]
            \node [anchor=west, rotate=90] (easl) at (0.01,0.65) {\small EASL};
            \node [anchor=west, rotate=90] (keasl) at (0.01,0.1) {\small KEASL};
            \node [anchor=west] (site1) at (0.15,-0.04) {\small (a)};
            \node [anchor=west] (site2) at (0.51,-0.04) {\small (b)};
            \node [anchor=west] (site3) at (0.83,-0.04) {\small (c)};
        \end{scope}

    \end{tikzpicture}
    \vspace*{-7mm}
    \caption{Depiction of the three test sites. The top row of images shows examples of EASL path planning through the environments, and the bottom row shows the KEASL planning. The columns are split into three pairs denoted by (a), (b), and (c), with each pair containing a binary cost map (left) and color-coded height map (right). The height is scaled from low to high elevation with blue to red coloring, respectively. In both the cost maps and the height maps, the paths are colored as a scale for robot velocity from 0m/s to the max allowable speed (2m/s) indicated from blue to red, respectively. The distance from the initial state (green rectangle in the center of each map) to every goal state (red rectangle at the end of each trajectory) is 25m.}
    \label{fig:EASL_multi_goal_1}
\end{figure*}

\begin{figure}[htb]
    \centering
    \resizebox{.9\columnwidth}{!}{\input{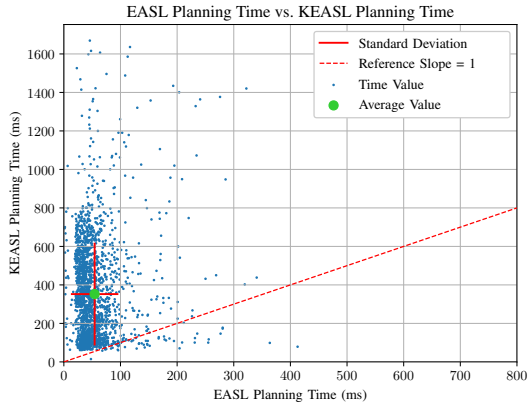}}
    \caption{Scatter plot depicting the EASL vs. KEASL time to generate a plan for a total of 2093 planned trajectories through the three test sites. The red line depicts a reference line for a slope of 1 for EASL vs. KEASL. Points lying below the line indicate a lower planning time by KEASL, and points above the line indicate a lower planning time by EASL.}
    \label{fig:planning_time_diff}
\end{figure}

The results from the physical experiments described in section \ref{sec:experimental-design} are shown in Figures \ref{fig:EASL_multi_goal_1}, \ref{fig:planning_time_diff}, \ref{fig:path_duration_diff}, and \ref{fig:adjusted_path_duration_diff} and Table \ref{fig:planning_metrics}.
Figure \ref{fig:EASL_multi_goal_1} shows plans from the current vehicle pose to a variety of sampled poses.
The top and bottom rows show plans generated by EASL and KEASL respectively, and the six columns illustrate samples of the three environments with cost map - height map pairs.
The differences between the EASL and KEASL plans in Figure \ref{fig:EASL_multi_goal_1}(a) show similar routes with a large disparity in planned velocities.
This environment is a mostly flat area with a tall, low sloped hill in the center.
The hill has lines of high grass lining a pathway to the top.
Here, there are similar path geometries because the shape of the explored search space is constrained by the corridor of obstacles.
Figure \ref{fig:EASL_multi_goal_1}(b) shows a similar variation in planned velocities for routes in the lower half of the map but less variation in the upper half where terrain geometry is less varied and obstacles are less dense.
This environment is a mostly flat area littered with downed trees and ditches, resulting in traversable terrain features that impose velocity constraints.
Figure \ref{fig:EASL_multi_goal_1}(c) shows a larger variation in both path geometries on the left half of the map and significant differences in the planned velocities through the narrow ravine on the right side of the map.
These illustrations show qualitative differences between the motions planned by EASL and KEASL from field experiments on a skid-steered mobile robot.  

\begin{figure}[htb]
    \centering
    \resizebox{.9\columnwidth}{!}{\input{figures/path_duration_diff.pgf}}
    \caption{Scatter plot showing the path duration of EASL vs. KEASL for a total of 2093 planned trajectories through the three test sites. The red line depicts a reference line for a slope of 1 for EASL vs. KEASL. Points lying below the line indicate a lower path duration by KEASL, and points above the line indicate a lower path duration by EASL.}
    \label{fig:path_duration_diff}
\end{figure}

\begin{figure}[htb]
    \centering
    \resizebox{.9\columnwidth}{!}{\input{figures/adjusted_path_duration_diff.pgf}}
    \caption{Scatter plot showing the path duration of EASL (when adjusted to satisfy terrain-dependent velocity constraints) vs. KEASL for a total of 2093 planned trajectories through the three test sites. The red line depicts a reference line for a slope of 1 for EASL vs. KEASL. Points lying below the line indicate a lower path duration by KEASL, and points above the line indicate a lower path duration by EASL.}
    \label{fig:adjusted_path_duration_diff}
\end{figure}

\begin{table}\small{
    \caption{Statistics for planning times and path durations of EASL, KEASL, and velocity-adjusted EASL (EASL$_{adj}$)}
    \label{fig:planning_metrics}
    \begin{center}
    \begin{tabular}{cccccc}
        \cline{2-6}
        & \multicolumn{2}{c|}{\emph{Planning Time (ms)}} & \multicolumn{3}{c}{\emph{Path Duration (s)}}\\
        & \multicolumn{1}{c}{\textbf{EASL}} & \multicolumn{1}{c|}{\textbf{KEASL}} & \multicolumn{1}{c}{\textbf{EASL}} & \multicolumn{1}{c}{\textbf{EASL$_{adj}$}} & \multicolumn{1}{c}{\textbf{KEASL}}\\
        \cline{1-6}
        \multicolumn{1}{c}{Mean} & \multicolumn{1}{|c}{54.35} & \multicolumn{1}{c|}{352.26} & \multicolumn{1}{c}{14.23} & \multicolumn{1}{c}{33.93} & \multicolumn{1}{c}{27.77}\\
        \multicolumn{1}{c}{Median} & \multicolumn{1}{|c}{43.60} & \multicolumn{1}{c|}{294.90} & \multicolumn{1}{c}{13.56} & \multicolumn{1}{c}{33.02} & \multicolumn{1}{c}{25.56}\\
        \multicolumn{1}{c}{$\sigma$} & \multicolumn{1}{|c}{40.22} & \multicolumn{1}{c|}{262.03} & \multicolumn{1}{c}{2.11} & \multicolumn{1}{c}{11.08} & \multicolumn{1}{c}{9.43}\\
        \hline
    \end{tabular}
    \end{center}
}
\end{table}

To more quantitatively assess the differences between EASL and KEASL, we generated 2093 planning problems using the sampling strategies illustrated in Figure \ref{fig:EASL_multi_goal_1} from a broader set of field experiment data and tracked the time required by both algorithms to plan motions, the difference in the duration of each planned motion, and whether the EASL plan violated any of the velocity constraints imposed by the constraints used by KEASL.  
Figure \ref{fig:planning_time_diff} shows the time required by each algorithm to plan as an x-y scatter plot where the independent axis is the EASL planning time and the dependent axis is the KEASL planning time.  
A dashed diagonal line indicates points where both algorithms take an even amount of time.  
In 97.76\% (2042 out of 2093) of plans, KEASL took longer to generate a plan than EASL for the sampled boundary states described in Section \ref{sec:experimental-design}.
This result is expected because of the added calculations performed by KEASL to perform kinodynamic analysis.
Interestingly, KEASL planned faster in 51 cases, despite the more significant per-edge cost of evaluation. 
This is due to KEASL eliminating parts of the search space that were deemed less or not viable because of the velocity constraints, which reduced the overall size of the sampled search space.
Figure \ref{fig:path_duration_diff} illustrates the difference in the planned durations of EASL and KEASL paths for identical motion planning problems.  
In 96.70\% (2024 out of 2093) of the plans, KEASL predicted a longer time of traversal than EASL.
This is expected because in most cases EASL assumed a constant acceleration to a max velocity for the duration of every path, whereas KEASL assumed slowdowns to obey the velocity limits and calculated the velocity along each trajectory.
Figure \ref{fig:adjusted_path_duration_diff} illustrates the duration of EASL generated solutions that were adjusted to satisfy KEASL's velocity constraints.
This process begins by estimating the roll and pitch values at each pose in the planned EASL trajectory.
Next, velocity acceleration constraints were updated based on these roll and pitch values and additional constraints imposed by the ground elevation map.
Finally, the adjusted duration of each EASL trajectory is calculated by rerunning the velocity planning procedure outlined in Algorithm \ref{alg:forward-backward-velocity-search}.
In 83.72\% (1800 out of 2093) of these plans, KEASL predicted a shorter time of traversal than EASL.
This result is also expected because KEASL heavily weighs duration in the path calculation, leading to shorter temporal trajectories.
It also shows the significant impact on the consideration of accurate velocity profiling along a trajectory to achieve faster paths.

The last metric tracked by these physical experiments was whether unadjusted EASL trajectories violated the velocity constraint imposed by KEASL.
Out of the 2093 planned paths, EASL violated velocity constraints in 2093 (100.0\%) of them, and roll limits in 376 (17.96\%) of them.
This means that EASL generated no trajectories that satisfied all of these constraints across the 2093 planning problems.
Although violation of the velocity constraints at the planner level is not indicative of a vehicle's inability to provide the slowdowns necessary to maintain a safe course, it could lead to catastrophic results if a controller was reliant on safe global plans.
KEASL is therefore more appropriate for scenarios where satisfaction of kinodynamic constraints by a global plan is a requirement for safe off-road navigation.
\section{Discussion}
\label{sec:discussion}

These real-world experimental results show that KEASL is able to provide more informed guidance about the time that it takes to plan through a non-flat environment when velocity constraints are imposed as a function of the terrain map.
The identified trade-off is the cost of computing the vehicle's attitude and elevation, the velocity acceleration constraints, and the velocity acceleration profile at each pose in the trajectory.
A robot is able to better exploit information about its environment when its cost function combines a traditional integrated cost-based metric with the duration of the planned motion.
Part of this is because the velocity acceleration constraints are a function of both the vehicle position and orientation, and such information cannot be encoded in a traditional 2-D cost map.% that is purely parameterized by position. 
Although the planning times are higher for KEASL than EASL, they remain well within the replanning rates for regional and/or global motion planning for off-road vehicles while providing much more informed guidance.
The sparsity and recombinant nature of state lattices and adaptivity of EASL is part of what enables KEASL to keep planning times reasonable, as sampling in the state space with a predefined edge library eliminates the burden of generating expressive sets of edges during search.
The results presented here use an assumed set of velocity acceleration constraints, and it should be noted that the differences in planning times, path durations, and percent of viable paths are all functions of the magnitude of these constraints.
Another significant outcome of this research is the integration of vehicle-specific models with velocity acceleration constraints into a state lattice-based motion planner.
Whereas previous versions of state lattice planners incorporated a more direct expression of vehicle velocity at nodes, KEASL is indexed by the direction in a manner that also enables it to express zero-velocity constraints.
Velocity and acceleration constraints are satisfied without having to ensure recombination in the graph with continuous velocity values.
This is achieved by uniquely computing each node based on the node's backpointer.
In environments where the difference between the path durations of EASL and KEASL are minimal, EASL should be applied so that the same plans can be made more efficiently. 
Learning when and where to apply the choice of EASL, KEASL, or other search spaces that may exhibit even better performance remains an open area of investigation and an exciting area of research.
\section{CONCLUSION}
\label{sec:conclusion}
In this paper, an extension of Efficiently Adaptive State Lattices called Kinodynmamic Efficiently Adaptive State Lattices is presented.
KEASL modified the representation of nodes in EASL and added steps for computing the attitude and elevation, velocity acceleration constraints, and velocity acceleration profiles that enabled search to produce better approximations of the time required to navigate routes in off-road environments.
Experimental results across 2093 planning problems illustrated the tradeoff of improved path duration evaluation by showing that KEASL requires more planning time than EASL in many cases, but also showed that the difference in planning time was not so significant that it prohibited use as a global planner on a field robot.
Future work will investigate the impact of more informed global guidance on navigation performance and whether the burden of compensating for paths that are uninformed by terrain imposed velocity constraints can be surmounted by sampling-based local planners.

\bibliography{citation.bib}

\begin{thebibliography}{10}

\bibitem{https://doi.org/10.1002/rob.20265}
Dave Ferguson, Thomas~M. Howard, and Maxim Likhachev.
\newblock Motion planning in urban environments.
\newblock {\em Journal of Field Robotics}, 25(11-12):939--960, 2008.

\bibitem{impugv}
Benned Hedegaard, Ethan Fahnestock, Jacob Arkin, Ashwin Menon, and Thomas~M.
  Howard.
\newblock Discrete optimization of adaptive state lattices for iterative motion
  planning on unmanned ground vehicles.
\newblock In {\em 2021 IEEE/RSJ International Conference on Intelligent Robots
  and Systems (IROS)}, pages 5764--5771, 2021.

\bibitem{rhmpc}
Thomas~M. Howard, Colin~J. Green, and Alonzo Kelly.
\newblock Receding horizon model-predictive control for mobile robot navigation
  of intricate paths.
\newblock In Andrew Howard, Karl Iagnemma, and Alonzo Kelly, editors, {\em
  Field and Service Robotics}, pages 69--78, Berlin, Heidelberg, 2010. Springer
  Berlin Heidelberg.

\bibitem{orttgwmr}
Thomas~M. Howard and Alonzo Kelly.
\newblock Optimal rough terrain trajectory generation for wheeled mobile
  robots.
\newblock {\em The International Journal of Robotics Research}, 26(2):141--166,
  2007.

\bibitem{844730}
James~J. Kuffner and Steven~M. LaValle.
\newblock Rrt-connect: An efficient approach to single-query path planning.
\newblock In {\em Proceedings 2000 ICRA. Millennium Conference. IEEE
  International Conference on Robotics and Automation. Symposia Proceedings
  (Cat. No.00CH37065)}, volume~2, pages 995--1001 vol.2, 2000.

\bibitem{4209319}
Nik~A. Melchior and Reid Simmons.
\newblock Particle rrt for path planning with uncertainty.
\newblock In {\em Proceedings 2007 IEEE International Conference on Robotics
  and Automation}, pages 1617--1624, 2007.

\bibitem{kmpslmp}
Mihail Pivtoraiko and Alonzo Kelly.
\newblock Kinodynamic motion planning with state lattice motion primitives.
\newblock In {\em 2011 IEEE/RSJ International Conference on Intelligent Robots
  and Systems}, pages 2172--2179, 2011.

\bibitem{dcmrmpsl}
Mihail Pivtoraiko, Ross~A. Knepper, and Alonzo Kelly.
\newblock Differentially constrained mobile robot motion planning in state
  lattices.
\newblock {\em Journal of Field Robotics}, 26(3):308--333, 2009.

\bibitem{sampocpf}
Venkataramanan Rajagopalan, Alonzo Kelly, and \c{C}etin Meri\c{c}li.
\newblock Slip-aware model predictive optimal control for path following.
\newblock In {\em 2016 IEEE International Conference on Robotics and Automation
  (ICRA)}, pages 4585--4590, 2016.

\bibitem{ompavtdt}
Zvi Shiller and J.C. Chen.
\newblock Optimal motion planning of autonomous vehicles in three dimensional
  terrains.
\newblock In {\em Proceedings., IEEE International Conference on Robotics and
  Automation}, pages 198--203 vol.1, 1990.

\bibitem{emop}
Jian Wen, Xuebo Zhang, Haiming Gao, Jing Yuan, and Yongchun Fang.
\newblock E³mop: Efficient motion planning based on heuristic-guided motion
  primitives pruning and path optimization with sparse-banded structure.
\newblock {\em IEEE Transactions on Automation Science and Engineering}, pages
  1--14, 2021.

\bibitem{mppi}
Grady Williams, Andrew Aldrich, and Evangelos Theodorou.
\newblock Model predictive path integral control using covariance variable
  importance sampling, 2015.

\end{thebibliography}
\bibliographystyle{plain}
\end{document}